\documentclass[letterpaper]{article} 
\usepackage{aaai24}  
\usepackage{times}  
\usepackage{helvet}  
\usepackage{courier}  
\usepackage[hyphens]{url}  
\usepackage{graphicx} 
\usepackage{natbib}  
\usepackage{caption} 
\usepackage{multirow}
\usepackage{enumitem}
\usepackage{amssymb}
\usepackage{booktabs}
\newcommand{\etal}{\textit{et al. }}
\newcommand{\ie}{\textit{i.e.}}

 \newcommand{\vs}{\textit{vs.}}
\usepackage[capitalize]{cleveref}
\crefname{section}{Sec.}{Secs.}
\Crefname{section}{Section}{Sections}
\Crefname{table}{Table}{Tables}
\crefname{table}{Tab.}{Tabs.}
\usepackage{color}

\usepackage{algorithm}
\usepackage{algorithmic}

\usepackage{newfloat}
\usepackage{listings}
\DeclareCaptionStyle{ruled}{labelfont=normalfont,labelsep=colon,strut=off} 
\floatstyle{ruled}
\newfloat{listing}{tb}{lst}{}
\floatname{listing}{Listing}
\title{TransGOP: Transformer-Based Gaze Object Prediction}
\author{
    Binglu Wang\textsuperscript{\rm 1, 2},
    Chenxi Guo\textsuperscript{\rm 1},
    Yang Jin\textsuperscript{\rm 1},
    Haisheng Xia\textsuperscript{\rm 3},
    Nian Liu\textsuperscript{\rm 4}\thanks{Nian Liu is the corresponding author.}}

\affiliations{
    \textsuperscript{\rm 1}Xi'an University of Architecture and Technology\\
    \textsuperscript{\rm 2}Beijing Institute of Technology\\
    \textsuperscript{\rm 3}University of Science and Technology of China\\
    \textsuperscript{\rm 4}Mohamed bin Zayed University of Artificial Intelligence\\

    \{wbl921129, guochenxix, jin91999\}@gmail.com, hsxia@ustc.edu.cn, liunian228@gmail.com
}

\usepackage{bibentry}

\begin{document}

\maketitle

\begin{abstract}
Gaze object prediction aims to predict the location and category of the object that is watched by a human. Previous gaze object prediction works use CNN-based object detectors to predict the object's location. However, we find that Transformer-based object detectors can predict more accurate object location for dense objects in retail scenarios. Moreover, the long-distance modeling capability of the Transformer can help to build relationships between the human head and the gaze object, which is important for the GOP task.
To this end,  this paper introduces Transformer into the fields of gaze object prediction and proposes an end-to-end Transformer-based gaze object prediction method named TransGOP. Specifically, TransGOP uses an off-the-shelf Transformer-based object detector to detect the location of objects and designs a Transformer-based gaze autoencoder in the gaze regressor to establish long-distance gaze relationships. Moreover, to improve gaze heatmap regression, we propose an object-to-gaze cross-attention mechanism to let the queries of the gaze autoencoder learn the global-memory position knowledge from the object detector. Finally, to make the whole framework end-to-end trained, we propose a Gaze Box loss to jointly optimize the object detector and gaze regressor by enhancing the gaze heatmap energy in the box of the gaze object. Extensive experiments on the GOO-Synth and GOO-Real datasets demonstrate that our TransGOP achieves state-of-the-art performance on all tracks, \ie, object detection, gaze estimation, and gaze object prediction. Our code will be available at https://github.com/chenxi-Guo/TransGOP.git.
\end{abstract}

\section{Introduction}

Predicting where a person is looking at a screen or an object has important applications in the real world, such as detecting goods of interest to people in retail scenarios, and fatigue detection is possible in autonomous driving, and in the medical field, it can help some patients with mobility or speech impairment to express their intentions~\cite{kleinke1986gaze,land2009looking, yu2022boundary,de2023using}.  

Previous research on gaze-related topics has primarily focused on gaze estimation (GE) task, which predicts heatmaps or points showing the location of human gaze objects. However, GE models fall short of identifying the exact location and category of the gaze object. Since the gaze object is closely linked to human behavior, it is important to identify the object due to its significant practical implications. 
Tomas \etal~\cite{tomas2021goo} proposed the gaze object prediction (GOP) task, which aims to predict both the location and category of the human gaze object. To facilitate research in the GOP task, they introduced the first dataset, the GOO dataset, which consists of images of people looking at different objects in retail scenarios. Compared to GE, the GOP task is more challenging as it requires richer information for model predictions.
Wang \etal~\cite{wang2022gatector} proposed GaTector, the first model for the GOP task, which combines a CNN-based object detector (YOLOv4~\cite{bochkovskiy2020yolov4}) with a CNN-based gaze prediction branch~\cite{chong2020detecting}. GaTector was trained and evaluated on the GOO dataset and achieved state-of-the-art performance on the GOP task.

\begin{figure}[!t]
    \centering
    \includegraphics[width=0.85\linewidth]{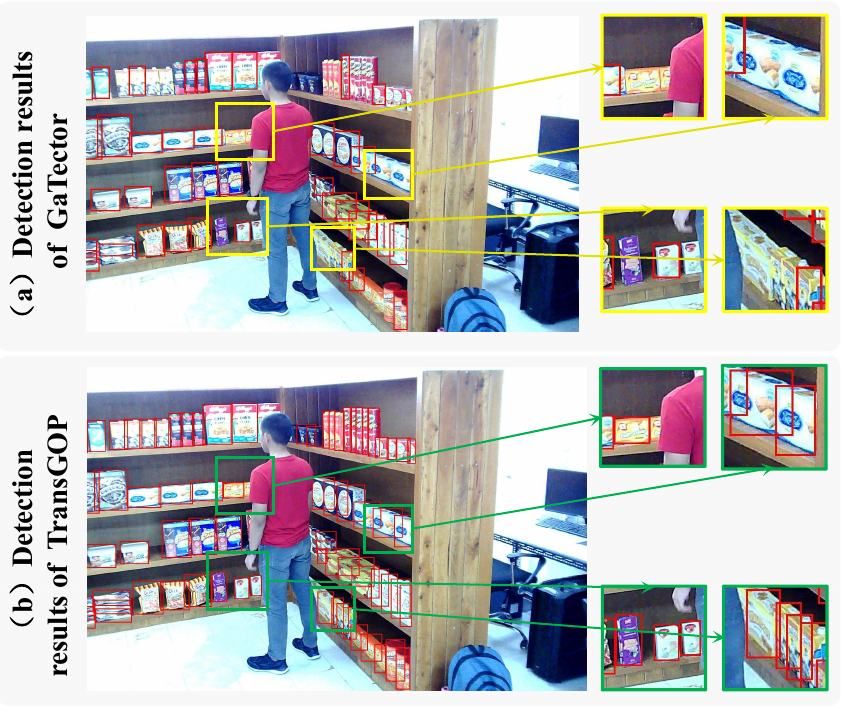}
    
    \caption{Object Detection results of GaTector~\cite{wang2022gatector} (a) and our TransGOP (b) when IoU threshold is 0.75. TransGOP predicts the object location more accurately than the GaTector, especially for the objects that are close to human or goods shelves.}
    \label{fig:intro}
\end{figure}

The GOP task is highly related to the object detection task, as both tasks aim to accurately localize objects in images. The accuracy of GOP is highly dependent on the accuracy of the object detector. CNN-based object detection has been extensively studied and achieved good performance scores~\cite{bochkovskiy2020yolov4, wang2023yolov7}.
However, in recent years, Transformer-based object detection methods have received increasing research attention~\cite{carion2020end,li2022exploring}. 
In this paper, we found that Transformer-based object detectors perform better than CNN-based object detectors in object-dense retail scenes (see Fig.~\ref{fig:intro}). 
Transformer-based object detectors are more effective at handling dense object scenes due to the attention mechanism, which provides them with long-range modeling capability.
There is no research on introducing Transformer-based methods into the GOP task. 
Compared to traditional methods, we believe that the capacity to capture long-range feature dependencies of Transformer methods could establish a better attention relationship between the human and the gaze object, further improving prediction accuracy.

In this paper, we introduce TransGOP, an end-to-end method for GOP tasks based on the Transformer architecture. TransGOP comprises two branches, an object detector and a gaze regressor, as shown in Fig.~\ref{fig:frame} (a). 
The object detector takes the entire image as input and detects the location and category of objects using an off-the-shelf Transformer-based object detector. 
The gaze regressor takes the head image and scene image as input and predicts the gaze heatmap. Specifically, in the gaze regressor, we design a Transformer-based gaze autoencoder to establish long-range dependencies of gaze-related features. 
To improve gaze heatmap regression, in the gaze autoencoder, we propose an object-to-gaze attention mechanism that enables the gaze autoencoder queries to learn global-memory position knowledge from the object detector (see in Fig.~\ref{fig:frame} (b)).

To facilitate end-to-end training of the model, we propose a Gaze Box loss that jointly optimizes the object detector and gaze regressor. 
As illustrated in Fig.~\ref{fig:frame} (c), our gaze box loss can further optimize the generation of gaze heatmaps through GT gaze boxes so that they can reflect object information.
Moreover,  the gradient backpropagation of gaze box loss is
propagated to the object detector backbone through scene features in the
gaze fusion module to achieve joint optimization.
Since the Transformer architecture can make the object detector end-to-end trained without any post-processing operation, our TransGOP would be the first end-to-end approach for the GOP task. Results on the GOO-Synth and GOO-Real datasets demonstrate that TransGOP outperforms existing state-of-the-art GOP methods by significant margins. 
To summarize, the contribution of this paper is fourfold:
\begin{itemize}
	\item We introduce the Transformer mechanism into the gaze object prediction task and propose an end-to-end model TransGOP.
	\item We propose an object-to-gaze cross-attention mechanism to establish the relationship between the object detector and gaze regressor.
	\item We propose a gaze box loss to jointly optimize the object detector and gaze regressor.
	\item Extensive experiments on the GOO-Synth and GOO-Real datasets show that TransGOP outperforms the state-of-the-art GOP methods.
\end{itemize}

\section{Related Works}
\label{sec:relat}

\subsection{Gaze Estimation}
Gaze estimation has important applications in many fields~\cite{kleinke1986gaze,land2009looking}, which aims to estimate where are people looking by taking eye or face images as input~\cite{yin2022nerf, balim2023efe}. 
Gaze point estimation~\cite{krafka2016eye,he2019device}, gaze following~\cite{judd2009learning,leifman2017learning,zhao2020learning,zhu2005eye}, and 3D gaze estimation~\cite{zhang2015appearance,zhang2017mpiigaze,cheng2018appearance,park2018deep} are sub-tasks of gaze estimation. The gaze-following task was first proposed by Recasens \etal~\cite{recasens2015they} who also released the dataset publicly. 
Chong \etal~\cite{chong2020detecting} proposed a cross-frame gaze-following model for video, which achieved a significant score metric.
Recently, some GE works~\cite{chengGazeEstimationUsing2021, guoMGTREndtoEndMutual, tuEndtoEndHumanGazeTargetDetection2022, yuGlanceandGazeVisionTransformer2021} introduced the transformer model. Tu \etal~\cite{tuEndtoEndHumanGazeTargetDetection2022} and Tonini \etal~\cite{tonini2023object} proposed transformer-based end-to-end GE model, which aims to estimate the human gaze heatmap but can not detect the bounding boxes and the categories of the gaze objects.

\subsection{Gaze Object Prediction}
Different from the GE task, the GOP task predicts not only the human gaze heatmap but also the location and category of the gaze object. 
The GOP task is first proposed by Tomas \etal~\cite{tomas2021goo}, who also contribute a novel dataset, \ie, the GOO dataset which consists of a large number of synthetic images (GOO-Synth dataset) and a smaller number of real images (GOO-Real dataset) of people gazing object in a retail environment. However, Tomas \etal do not propose a model to resolve the GOP problem. 
Afterward, Wang \etal~\cite{wang2022gatector} propose the first unified framework GaTector for the GOP task which utilizes a CNN-based object detector~\cite{bochkovskiy2020yolov4} to detect the objects and design another CNN-based gaze prediction branch~\cite{chong2020detecting} to predict the gaze heatmap. To further improve the performance of gaze estimation, GaTector proposed an energy aggregation loss to supervise the range of gaze heatmaps.

In this paper, we want to propose a Transformer-based GOP model due to the long-distance modeling capability of the Transformer can help to build human-object relationships, which can improve the performance of the GOP task.

\begin{figure*}[!t]
  \centering
  \includegraphics[scale=0.89]{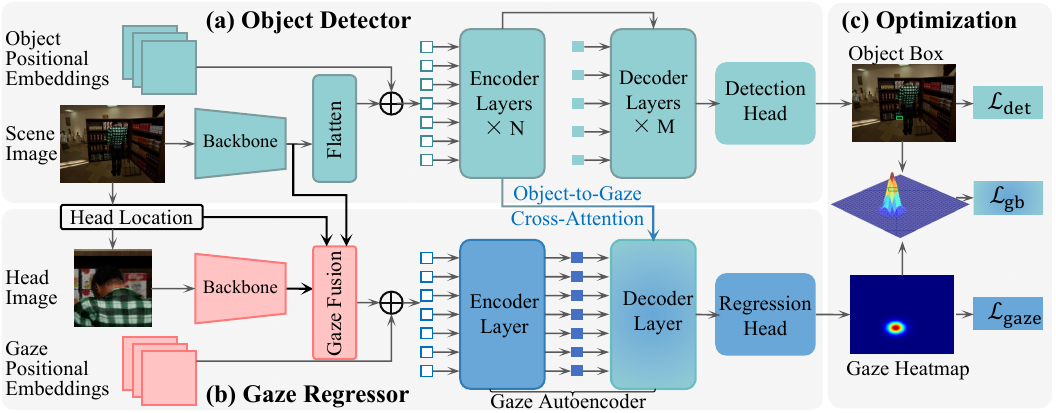}
  \caption{Overview framework of our TransGOP method. (a) The object detector in TransGOP is the of-the-shelf Transformer-based object detection method that detects object location and category. (b) The gaze regressor feeds the fused feature into the Transformer-based gaze autoencoder to predict the gaze heatmap. (c) The optimization of TransGOP consists of three parts: the object detection loss $\mathcal{L}_{\rm det}$ for optimizing the object detector, the gaze regression loss $\mathcal{L}_{\rm gaze}$ for optimizing the gaze regressor, and the gaze box loss $\mathcal{L}_{\rm gb}$ to jointly optimize the object detector and gaze regressor.}
  \label{fig:frame}
\end{figure*}

\subsection{Object Detection}
Object detection (OD) is a fundamental task in computer vision. CNN-based object detectors can be classified into anchor-based~\cite{he2015spatial,girshick2015fast, law2018cornernet,duan2019centernet,bochkovskiy2020yolov4} and anchor-free~\cite{huang2015densebox,yang2020circlenet,zhou2019bottom,tian2022fully,kong2020foveabox} methods. 

Recently, Transformer has also been widely applied to resolve the object detection task~\cite{vaswani2017attention,dai2021up,fang2021you,liu2022swin,li2022exploring}. DETR~\cite{carion2020end} is the first Transformer-based method that treats object detection as a set sequence prediction problem. Many subsequent DETR series methods~\cite{zhu2020deformable,meng2021conditional,wang2022anchor,liu2022dab,li2022dn, zheng2023less} are trying to resolve the problem of DETR about slow convergence and low precision. DINO~\cite{zhang2022dino} achieves better performance and faster speed of convergence than previous DETR-like models by using contrastive denoising training methods and excellent query design strategy. 

In this paper, we find that Transformer-based object detectors can predict more accurately than CNN-based object detectors, especially in dense object retail scenarios, so we want to propose a Transformer-based method for the GOP task to better utilize this advantage.

\section{Method}
\label{sec:method}

Given a scene image and a head image, our goal is to predict the category, bounding box, and gaze heatmap for the human gaze object. In this section, we first present the overall framework of TransGOP, then we introduce the object detector and the gaze regressor in detail. Finally, we give a detailed introduction to the proposed gaze box loss function.

\subsection{Overview}
\label{sec:overall}
As illustrated in Fig.~\ref{fig:frame}, the proposed TransGOP consists of an object detector and a gaze regressor. The object detector is a Transformer-based object detection method, which takes the whole scene image as input and predicts categories and locations for all objects. The gaze regressor has a Transformer-based gaze autoencoder, which takes the head image, scene image, and head location map as input and generates queries with gaze information, which will be regressed to the gaze heatmap by the regression head. In inference, the gaze object is determined by the value of gaze heatmap energy in the predicted object boxes. The overall loss function during the training process is defined as:
\begin{equation}
  \label{eq:loss}
  \mathcal{L} = \mathcal{L}_{\rm det} + \alpha\mathcal{L}_{\rm gaze} +\beta\mathcal{L}_{\rm  gb},
\end{equation}
where $\alpha$ and $\beta$ are the weights of the gaze heatmap loss and the gaze box loss, respectively.

For the Transformer-based object detector, we directly use an existing method DINO~\cite{zhang2022dino}. The object detector takes the whole image as input and aims to predict the object category and bounding box. The loss of the object detector is denoted by $\mathcal{L}_{\rm det}$.

This paper proposes a novel Transformer-based gaze regressor to predict the gaze heatmap. The gaze regressor first extracts the head feature of the person in the image, then the fused feature the head feature, scene feature, and head location map as the input of the Transformer-based gaze autoencoder. The gaze heatmap is optimized by $\mathcal{L}_{\rm gaze}$.

Finally, we propose a new gaze box loss function $\mathcal{L}_{\rm  gb}$ to jointly optimize the object detector and the gaze regressor, and make the whole framework be trained end-to-end.

\subsection{Transformer-based Object Detector}

As we illustrate in Fig.~\ref{fig:frame} (a), a DETR-like object detector usually consists of a backbone to extract semantic features, multiple layers of Transformer encoders, multiple layers of Transformer decoders, and a prediction head. In this paper, we use DINO~\cite{zhang2022dino} as our object detector, which achieves remarkable results with a convergence speed similar to previous CNN-based methods, surpassing the accuracy of other DETR series models.

The loss of our object detector $\mathcal{L}_{\rm det}$ consists of a classification loss and a box regression loss. Following the setting of DINO, we employ focal loss~\cite{lin2017focal} as the classification loss to address the imbalanced positive and negative samples and use L1 and GIoU loss~\cite{rezatofighi2019generalized} to supervise the regression of the predicted box. 

It is notable that the key-value pairs $\{ \mathbf{K}_{\rm det}^{\rm en}, \mathbf{V}_{\rm det}^{\rm en} \in \mathbb{R}^{D\times M}\}$ from the object encoder are fed to the decoder layer of the gaze autoencoder to make the query in the gaze regressor decoder can perceive objects in the scene. Where $D=256$ is the hidden size of tokens, and $M=1045$ is the sum of spatial scale for multi-scale features in DINO. Moreover, our proposed \textit{TransGOP can use any DETR-like method as the object detector, as long as it can provide keys and values from the encoder.}

\begin{figure}[!t]
  \centering
  \includegraphics[width=1\linewidth]{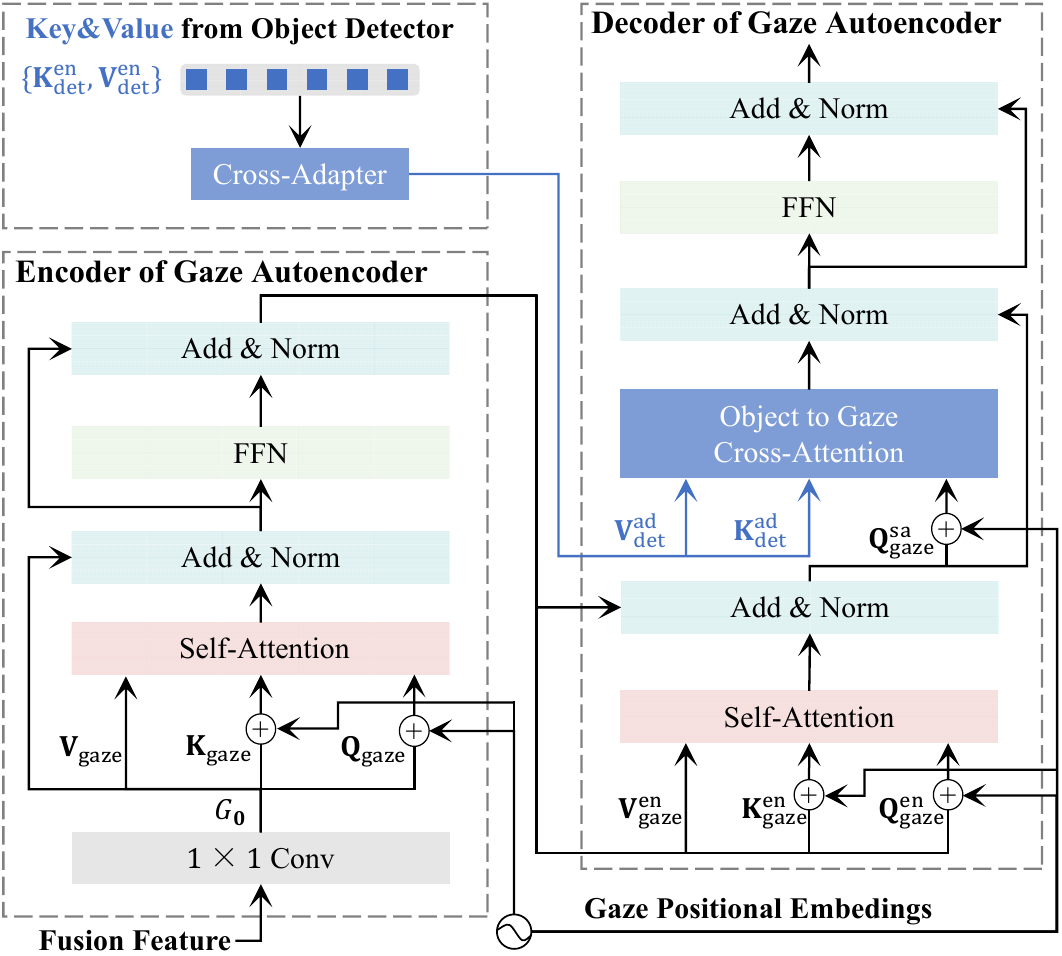}
  \caption{Details of the Transformer-based gaze autoencoder and the object-to-gaze cross-attention in the gaze regressor.}
  \label{fig:corss}
\end{figure}

\subsection{Transformer-based Gaze Regressor}
\label{sec:regressor}
The gaze regressor aims to predict a gaze heatmap for the human gaze object. As illustrated in Fig.~\ref{fig:frame} (b), we use the head image, scene image, and head location map as the input of the gaze regressor. The gaze regressor consists of a gaze backbone, a gaze fusion module, a single-layer Transformer-based gaze autoencoder, and a gaze predictor.

\noindent\textbf{Input.}
As illustrated in the left part of Fig.~\ref{fig:frame} (b), the scene image is resized into the size of ${3\times H_0\times W_0}$, where $H_0, W_0=224$. The head location map is a binary map with the same size as the scene image, where the value in the gaze box is 1 and 0 otherwise. The head image is obtained by cropping the scene image according to the head location. 

\noindent\textbf{Gaze feature fusion.}
In this paper, we use the gaze fusion module proposed in the GaTector~\cite{wang2022gatector} to fuse features from the head image and scene image. The scene image and head image are fed into two independent Resnet50 backbones to extract the salient feature of the scene and head direction feature, respectively. The head location map is fed into five convolutional layers to generate head location features that have the same spatial resolution as image features~\cite{wang2022gatector}. Then, the head features and scene features are fused with the help of the head location features. Specifically, the gaze feature module first stacks the head feature with the head location map and feeds it into a linear layer to generate an attention map that encodes directional cues. By computing the inner product between the attention map and the scene feature map, a fused gaze feature $\mathbf{F}_{\rm fuse}\in \mathbb{R}^{C\times H\times W}$ is generated, which can enhance the region that is highly related to the gaze behavior. Typical values we set are $C=256, H, W=15$.

\noindent\textbf{Encoder.}
Different from GaTector which directly predicts the gaze heatmap from the fused feature, we propose a Transformer-based gaze autoencoder to build the long-range relationship for better gaze heatmap prediction results. 
As illustrated in Fig.~\ref{fig:corss}, a $1\times1$ convolution operation is used to reduce the channel dimension of the fused gaze feature $\mathbf{F}_{\rm fuse}$ from $C$ to the hidden size $D$, and get a new feature map $\mathbf{G}_0\in \mathbb{R}^{D\times H\times W}$. The spatial dimensions of $\mathbf{G}_0$ are collapsed into one dimension and result in a $D\times HW$ feature map. Then, we generate queries, keys, and values $\{\mathbf{Q}_{\rm gaze}, \mathbf{K}_{\rm gaze}, \mathbf{V}_{\rm gaze} \in \mathbb{R}^{D\times HW}\}$ for the encoder, which consists of a multi-head self-attention module and a feed-forward network (FFN). Gaze positional embeddings are also added to the input of each attention layer.

\noindent\textbf{Decoder and Object-to-gaze cross-attention.}
The decoder layer of the gaze autoencoder mainly consists of a self-attention block and an object-to-gaze cross-attention block. As shown in Fig.~\ref{fig:corss}, we use the encoded queries, keys, and values $\{\mathbf{Q}_{\rm gaze}^{\rm en}, \mathbf{K}_{\rm gaze}^{\rm en}, \mathbf{V}_{\rm gaze}^{\rm en} \in \mathbb{R}^{D\times HW}\}$ as the input of the self-attention block, gaze positional embeddings are also added to queries and keys of each self-attention layer.

To improve gaze heatmap regression, this paper proposes an object-to-gaze cross-attention mechanism to make gaze autoencoder learn more accurate position information about the gaze object. As illustrated in Fig.~\ref{fig:corss}, we first fed key-value pairs from the encoder of object detector, \ie, $\{\mathbf{K}_{\rm det}^{\rm en}, \mathbf{V}_{\rm det}^{\rm en}\}$, into a cross-adapter module to optimizes the keys and values $\{\mathbf{K}_{\rm det}^{\rm ad}, \mathbf{V}_{\rm det}^{\rm ad}\}$ to provide more specific information about the gaze object. $\{\mathbf{K}_{\rm det}^{\rm ad}, \mathbf{V}_{\rm det}^{\rm ad}\}$ are used as the input of the object-to-gaze cross-attention module in the gaze decoder layer. Moreover, the output queries from the self-attention module in the decoder---$\mathbf{Q}_{\rm gaze}^{\rm sa}$---are used as the input queries of the cross-attention module, and gaze positional embeddings are also added to $\mathbf{Q}_{\rm gaze}^{\rm sa}$. We use the object-to-gaze cross-attention mechanism to make $\mathbf{Q}_{\rm gaze}^{\rm sa}$ learn global-memory position knowledge from the object detector, so can help to predict a more accurate gaze heatmap for the gaze object. 

\noindent\textbf{Gaze prediction.}
After the gaze autoencoder, we fed the decoded features into a regression head to predict the gaze heatmap. In this paper, we use the same regression head as the GaTector~\cite{wang2022gatector} and generate ground truth gaze heatmaps $\mathbf{T}\in \mathbb{R}^{H_{\rm T} \times W_{\rm T}}$ by Gaussian blurred gaze points $(p_{x}, p_{y})$:
\begin{equation}
      \widetilde{\mathbf{T}} = \frac{1}{2 \pi \sigma_{x} \sigma_{y} } \exp \left[-\frac{1}{2}\left(\frac{\left(x-q_{x}\right)^{2}}{\sigma_{x}^{2}}+\frac{\left (y-q_{y}\right)^{2}}{\sigma_{y}^{2}}\right)\right],
  \label{eqGaussian}
\end{equation}
\begin{equation}
  \mathbf{T}_{i,j} = \frac{\widetilde{\mathbf{T}}_{i,j}}{{\rm max}(\widetilde{\mathbf{T}})},
\label{TNorm}
\end{equation}
where $\widetilde{\mathbf{T}}$ is the Gaussian-blurred heatmap and ${\rm max}(\widetilde{\mathbf{T}})$ is its maximum value. $\sigma_{x}$ and $\sigma_{y}$ indicate the standard deviation, which we follow Chong \etal~\cite{chong2020detecting} to set $\sigma_{x} = 3$, $\sigma_{y} = 3$. $H_{\rm T}$ and $W_{\rm T}$ represent the height and width of the heatmap.

Suppose the predicted gaze heatmaps is $\mathbf{M}\in \mathbb{R}^{H_{\rm T} \times W_{\rm T}}$, we use the mean square between $\mathbf{M}$ and $\mathbf{T}$ calculate the gaze heatmap loss:
\begin{equation}
    \mathcal{L}_{\rm gaze} = \frac{1}{H_{\rm T} \times W_{\rm T}} \sum_{i=1}^{H_{\rm T}} \sum_{j=1}^{W_{\rm T}} (\textbf{M}_{i,j} - \textbf{T}_{i,j})^{2}.
    \label{eqLgaze}
\end{equation}

\begin{figure}[!t]
  \centering
  \includegraphics[scale=0.8]{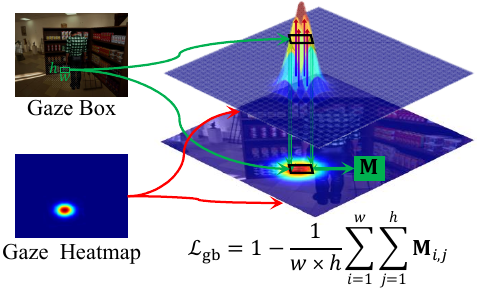}
  \caption{Illustration of the gaze box loss.}
  \label{fig:loss}
\end{figure}

\subsection{Gaze Box loss}
\label{sec:loss}
Since DETR-like object detectors can achieve end-to-end training without any extra post-processing operations, we want to make our proposed Transformer-based gaze regressor can be trained together with the DETR-like object detector to obtain an end-to-end framework for the GOP task. In this paper, we propose a gaze box loss to jointly optimize the object detector and gaze regressor. 

In previous work~\cite{wang2022gatector}, an energy aggregation loss was proposed to optimize the gaze heatmap regression process by supervising the overall distribution of the heatmap to concentrate in the gaze box. 
However, we find that the actual predicted gaze heatmap area is usually much larger than the area of the gaze box, indicating that the energy aggregation loss has a very limited role in optimizing the gaze regressor.
Moreover, because the gradient backpropagation of the gaze heatmap will affect the object detector backbone through the scene features in the gaze fusion, Table~\ref{tab:component} illustrates that the energy aggregation loss could cause a decrease in object detection performance.

To this end, we want to propose a novel loss to provide a positive effect for both the object detector and gaze regressor. As shown in Fig.~\ref{fig:loss}, the proposed gaze box loss aims to focus on improving the heatmap energy in the gaze box:
\begin{equation}
    \mathcal{L}_{\rm gb} = {1} - \frac{1}{h \times w} \sum_{i={x}_{1}}^{{x}_{2}} \sum_{j={y}_{1}}^{{y}_{2}} \textbf{M}_{i,j}.
    \label{eqenergy}
\end{equation}
where ${h},{w}$ are the height and width of the ground truth gaze box. The gaze box loss aims to maximize the high-energy value part of the predicted gaze heatmap distributed within the gaze box, to accurately reflect object position information. 
Experimental results in Table~\ref{tab:component} demonstrate that our gaze box loss has a positive effect on both object detection and gaze heatmap regression process.

\section{Experiments}
\label{sec:experiments}

\subsection{Datasets \& Settings}
\label{sec:settings}
All experiments were conducted on GOO-Synth and GOO-Real datasets~\cite{tomas2021goo}.
TransGOP is trained for 50 epochs, with an initial learning rate of ${1e-4}$, and the learning rate is reduced by 0.94 times every 5 epochs. 
We use AdamW as our optimizer.
For the gaze autoencoder, we set the hidden size to 256 and employ 200 decoder queries.
In Eq.\ref{eq:loss}, the loss weight $\alpha$ is 1000 and $\beta$ is 10. 
The input size of the image is set to $224\times224$ and the predicted heatmap size is $64\times 64$. 
All experiments are implemented based on the PyTorch and one GeForce RTX 3090Ti GPU.

\noindent\textbf{Metrics}
We use the Average Precision (AP) to evaluate the object detection performance.
For gaze estimation, we adopt commonly used metrics including AUC, L2 distance error (Dist.), and angle error (Ang.). 

For the gaze object prediction, we utilize mSoC\footnote{\noindent The mSoC metric is proposed by Wang~\etal in the latest arXiv version.URL: \url{https://arxiv.org/abs/2112.03549}. \\The code can be available at \url{ https://github.com/CodeMonsterPHD/GaTector-A-Unified-Framework-for-Gaze-Object-Prediction.git}. }
metric that can measure differentiation even when the predicted box and ground truth box do not overlap with each other.

\begin{table}[!t]
  \centering
  
  \resizebox{0.88\linewidth}{!}{
    \begin{tabular}{lcccc}
    \toprule
    Methods& mSoC & mSoC$_{50}$ & mSoC$_{75}$ & mSoC$_{95}$ \\
    \midrule
    GaTector 
    &    67.9   &    98.1   &   86.2    & 0.1 \\
    
    TransGOP 
    &\textbf{92.8}       &\textbf{99.0}       &\textbf{98.5}       &\textbf{51.9}  \\
    \bottomrule
    \end{tabular}%
    }
    \caption{Gaze object prediction results on GOO-Synth.}
  \label{tab:GOP_synth}%
\end{table}%

\begin{table}[!t]
  \centering
  
\resizebox{0.88\linewidth}{!}{
    \begin{tabular}{lcccc}
    \toprule
    Methods &  mSoC  & mSoC$_{50}$ & mSoC$_{75}$ & mSOC$_{95}$\\
    \midrule
    \multicolumn{5}{l}{No Pre-train}\\
    \midrule
    GaTector  &   62.4    &    95.1   &    73.5   &  0.2        \\
    TransGOP  &82.6       &98.3       &93.5       &15.3                \\
    \midrule
    \multicolumn{5}{l}{Pre-trained on GOO-Synth}\\
    \midrule
    GaTector  & 71.2    & 97.5    & 88.7    & 0.4      \\
    TransGOP  &\textbf{89.0}       &\textbf{98.9}       &\textbf{97.5}       &\textbf{33.2}               \\
    \bottomrule
    \end{tabular}%
    }
    \caption{Gaze object prediction results on GOO-Real.}
  \label{tab:GE_GOP_real}%
\end{table}%

\subsection{Comparison to State-of-the-Art}

\noindent\textbf{Gaze object prediction.}
Table ~\ref{tab:GOP_synth} and Table ~\ref{tab:GE_GOP_real} show
the performance comparison with the GaTector~\cite{wang2022gatector} on the
GOO-Synth and GOO-Real datasets respectively.

As shown in Table~\ref{tab:GOP_synth}, TransGOP achieves better results, which is 24.9\% mSoC higher than Gatector (92.8\% \vs 67.9\%). The performance of GaTector drops sharply with increasingly strict mSoC constraints, while TransGOP can achieve comparable results. 
On the GOO-Real dataset in Table~\ref{tab:GE_GOP_real}, TransGOP can achieve a comparable performance of 82.6\% mSoC without pre-train. GaTector requires pre-training to achieve better performance (71.2\% mSoC) on the GOO-Real dataset.
This demonstrates the powerful modeling and feature extraction capabilities rooted in the Transformer, which enable our model to exhibit better GOP performance and generalization capability.

\begin{table}[!t]
  \centering
  
  \resizebox{0.91\linewidth}{!}{
    \begin{tabular}{llcccc}
    \toprule
    &Methods & AP    & AP$_{50}$ & AP$_{75}$ & AP$_{95}$  \\
    
    \midrule
    \multirow{6}{*}{\rotatebox{90}{GOO-Synth}}&YOLOv4 & 46.6 & 88.1 & 46.8 & 0.1   \\
    &YOLOv7 & 54.4 & 98.3 & 54.4 & 0.2   \\
    &GaTector & 56.8 & 95.3 & 62.5 & 0.1   \\
    &Deformable DETR & 81.0 & 98.3 & 92.8 & 14.9  \\
    &DINO & 83.7 & 98.5 & 93.9 & 21.7  \\
    &TransGOP &\textbf{87.6}       &\textbf{99.0}       &\textbf{97.3}       &\textbf{25.5}         \\
    
    \midrule
    \multirow{6}{*}{\rotatebox{90}{GOO-Real}}&YOLOv4   & 43.7 & 84.0 & 43.6 & 0.1 \\
    &YOLOv7   & 57.3 & 96.4 & 63.2 & 0.3  \\
    &GaTector   & 52.2 & 91.9 & 55.3 & 0.1 \\
    &Deformable DETR  & 81.3 & 96.0 & 93.4 & 11.3 \\
    &DINO &  82.8 & 98.7 & 94.6 & 13.5 \\
    &TransGOP   &\textbf{84.1}       &\textbf{98.8}       &\textbf{94.7}       &\textbf{20.1}  \\
    \bottomrule
    \end{tabular}%
    }
    \caption{Object detection results on GOO-Synth and GOO-Real datasets.}
  \label{tab:object_detection}%
\end{table}%

\noindent\textbf{Object detection.} Comparison with CNN-based object detectors YOLOv4~\cite{bochkovskiy2020yolov4}, YOLOv7~\cite{wang2023yolov7}, GaTector~\cite{wang2022gatector} and Transformer-based object detectors Deformable DETR~\cite{zhu2020deformable}, DINO~\cite{zhang2022dino} on the GOO-Synth dataset and GOO-Real datasets in Table~\ref{tab:object_detection}. 
The global modeling capability of the Transformer makes Transformer-based methods have better performance in dense object scenes, such as the GOO-Synth dataset and GOO-Real datasets, than CNN-based methods. 
TransGOP also achieves good object detection performance, and although our object detector is based on DINO, through the joint optimization capability of gaze box loss, TransGOP object detection performance exceeds DINO.

\noindent\textbf{Gaze estimation.} 
Comparison with SOTA GE methods~\cite{recasens2017following,lian2018believe,chong2020detecting,wang2022gatector} on the GOO-Synth dataset is summarized in Table~\ref{gaze_estimation}.
Our TransGOP achieves current SOTA performance with AUC (0.963) and angular error ($13.30^{\circ}$). This is attributed to the capability of our proposed gaze autoencoder to establish long-distance gaze relationships, while the object-to-gaze cross-attention mechanism enables it to learn global-memory position knowledge, thereby enhancing gaze estimation performance.
The L2 distance error of TransGOP slightly lags behind GaTector by 0.006 (0.079 \textit{vs.} 0.073). Because our gaze box loss makes the predicted gaze heatmap prefer to reflect the object position information, this results in a slightly worse L2 distance error calculated using gaze points.
Meanwhile, Our TransGOP also outperforms GaTector on the GOO-Real dataset in Table~\ref{tab:GE_real}.

\begin{table}[!t]
  \centering
  
  \resizebox{0.634\linewidth}{!}{
    \begin{tabular}{lccc}
    \toprule
    Methods & AUC↑  & Dist.↓ & Ang.↓ \\
    \midrule
    Random & 0.497 & 0.454  & 77.0  \\
    Recasens  & 0.929 & 0.162  & 33.0  \\
    Lian  & 0.954 & 0.107  & 19.7  \\
    Chong  & 0.952 & 0.075  & 15.1  \\
    GaTector  & 0.957 & \textbf{0.073}  & 14.9  \\
    TransGOP  & \textbf{0.963} & 0.079  & \textbf{13.3}  \\
    \bottomrule
    \end{tabular}%
    }
    \caption{Gaze estimation results on GOO-Synth.}
  \label{gaze_estimation}%
\end{table}%

\begin{table}[!t]
  \centering
  
    \resizebox{0.634\linewidth}{!}{
    \begin{tabular}{lccc}
    \toprule
    Methods &  AUC↑  & Dist.↓ & Ang.↓  \\
    \midrule
    \multicolumn{4}{l}{No Pre-train}\\
    \midrule
    GaTector  &   0.927    &   0.196    &    39.50   \\
    TransGOP  & 0.947      &0.097       &16.73         \\
    \midrule
    \multicolumn{4}{l}{Pre-trained on GOO-Synth}\\
    \midrule
    GaTector  & 0.940    & 0.087   & 14.79   \\
    TransGOP  &\textbf{0.957}       &\textbf{0.081}       &\textbf{14.71}         \\
    \bottomrule
    \end{tabular}%
    }
    \caption{Gaze estimation results on GOO-Real.}
  \label{tab:GE_real}%
\end{table}%

\begin{table*}[!t]
  \centering

\resizebox{0.98\linewidth}{!}{
    \begin{tabular}{c|cc|c|cccc|cccc|ccc}
    \toprule
        \multirow{2}{*}{Baseline}& \multicolumn{2}{c|}{Loss}& \multirow{2}{*}{GA}& \multicolumn{4}{c|}{Gaze Object Prediction} &\multicolumn{4}{c|}{Object Detection}&\multicolumn{3}{c}{Gaze Estimation}\\
          & $\mathcal{L}_{\rm eng}$ & $\mathcal{L}_{\rm gb}$ &  & mSoC & mSoC$_{50}$ & mSoC$_{75}$ & mSOC$_{95}$& AP & AP$_{50}$ & AP$_{75}$ & AP$_{95}$& AUC↑ & Dist.↓ & Ang.↓ \\
    \midrule
    $\surd$ &       &       &       & 84.9  & 98.5  & 93.2  & 27.3  & 83.7  & 98.5  & 93.9  & 21.7& 0.949 & 0.095  & 17.90   \\
    $\surd$ & $\surd$ &       &       & 85.4  & 98.5  & 95.1  & 28.7  & 81.2  & 98.3  & 92.7  & 20.2& 0.952 & 0.091  & 16.60   \\
    $\surd$ &       & $\surd$ &       & 90.4  & 99.0  & 97.5  & 41.2  & 84.1  & 98.7  & 95.1  & 22.7& 0.958 & 0.089  & 15.93   \\
    $\surd$ &       &       & $\surd$ & 89.8  & 98.9  & 97.2  & 38.6  & 85.0  & 98.9  & 96.2  & 23.2& 0.955 & 0.092  & 14.31   \\
    $\surd$ &       & $\surd$ & $\surd$ &\textbf{92.8}       &\textbf{99.0}       &\textbf{98.5}       &\textbf{51.9}       &\textbf{87.6}       &\textbf{99.0}       &\textbf{97.3}       &\textbf{25.5}       &\textbf{0.963}       &\textbf{0.079}       &\textbf{13.30}   \\
    \bottomrule
    \end{tabular}%
    }
      \caption{Ablation comparison about each component on the GOO-Synth dataset.}
  \label{tab:component}%
\end{table*}%

\begin{table*}[!t]
  \centering

  \resizebox{0.9\linewidth}{!}{
    \begin{tabular}{c|cccc|cccc|ccc}
    \toprule
    \multirow{2}{*}{Setups}& \multicolumn{4}{c|}{Gaze Object Prediction} &\multicolumn{4}{c|}{Object Detection}&\multicolumn{3}{c}{Gaze Estimation}\\
     &mSoC & mSoC$_{50}$ & mSoC$_{75}$ & mSoC$_{95}$& AP & AP$_{50}$ & AP$_{75}$ & AP$_{95}$& AUC↑ & Dist.↓ & Ang.↓ \\
    \toprule
    w/o cross-adapter & 90.8  & 99.0  & 97.8  & 41.8  & 85.8  & 98.0  & 96.5  & 23.8& 0.960  & 0.084  & 14.20    \\
    w/ cross-adapter &\textbf{92.8}       &\textbf{99.0}       &\textbf{98.5}       &\textbf{51.9}       &\textbf{87.6}       &\textbf{99.0}       &\textbf{97.3}       &\textbf{25.5}       &\textbf{0.963}       &\textbf{0.079}       &\textbf{13.30}   \\
    \bottomrule
    \end{tabular}
    }
    \caption{Ablation study about cross-adapter on the GOO-Synth dataset.}
  \label{tab:adapter}%
\end{table*}%

\begin{table*}[!t]
  \centering
  
  \resizebox{0.869\linewidth}{!}{
    \begin{tabular}{c|cccc|cccc|ccc}
    \toprule
    \multirow{2}{*}{Setups}& \multicolumn{4}{c|}{Gaze Object Prediction} &\multicolumn{4}{c|}{Object Detection}&\multicolumn{3}{c}{Gaze Estimation}\\
    &mSoC & mSoC$_{50}$ & mSoC$_{75}$ & mSoC$_{95}$& AP & AP$_{50}$ & AP$_{75}$ & AP$_{95}$& AUC↑ & Dist.↓ & Ang.↓ \\
    \midrule
    gaze-to-object & 79.4  & 93.1  & 89.3  & 13.2  & 70.0  & 94.4  & 79.4  & 4.2& 0.860  & 0.173  & 23.43   \\
    object-to-gaze &\textbf{92.8}       &\textbf{99.0}       &\textbf{98.5}       &\textbf{51.9}       &\textbf{87.6}       &\textbf{99.0}       &\textbf{97.3}       &\textbf{25.5}       &\textbf{0.963}       &\textbf{0.079}       &\textbf{13.30}   \\
    \bottomrule
    \end{tabular}
    }
    \caption{Comparison of the different cross-attention mechanisms on the GOO-Synth dataset.}
  \label{tab:ge-to-od}%
\end{table*}%

\subsection{Ablation Studies and Model Analysis}
\label{ablation}
\noindent\textbf{Ablation study about each component.} We conducted ablation studies in Table~\ref{tab:component} to analyze the effectiveness of our proposed \textbf{gaze autoencoder} (GA) and \textbf{gaze box loss} ($\mathcal{L}_{\rm gb}$) on the GOO-Synth dataset. 
In the first row of Table~\ref{tab:component}, we build a strong baseline by combining DINO with a CNN-based gaze regressor~\cite{chong2020detecting} for comparison.

We first compare our $\mathcal{L}_{\rm gb}$ to the energy aggregation loss ($\mathcal{L}_{\rm eng}$) proposed in Gatector~\cite{wang2022gatector}. Based on the results, $\mathcal{L}_{\rm eng}$ can enhance GE and GOP performance of the baseline but leads to a 2.5\% decrease in object detection AP (81.2\% \vs 83.7\%). 
However, our $\mathcal{L}_{\rm gb}$ can jointly optimize performance gains for both GE and OD. 
The baseline with the gaze autoencoder (in Table~\ref{tab:component} fourth line) achieves a 4.9\% mSoC improvement in GOP performance (89.8\% \vs 84.9\%), which demonstrates that the gaze autoencoder can establish long-distance gaze relationships, thus predict the more accurate gaze object. 
Our complete model can achieve 92.8\% mSoC. From the experimental results, our proposed methods in TransGOP further improve the performance significantly.

\noindent\textbf{Ablation study about cross-adapter.} Table~\ref{tab:adapter} reports the effectiveness of the cross-adapter in the gaze autoencoder. 
The results show that the cross-adapter can improve the performance of GOP (92.9\% vs 90.8\%), which is mainly because the cross-adapter can optimize the task conversion from the key-value pair of the object detector to the gaze autoencoder.

\noindent\textbf{Comparative between different cross-attention mechanisms.} In Table~\ref{tab:ge-to-od}, we compare the effectiveness of different cross-attention mechanisms, \ie, the gaze-to-object and object-to-gaze mechanism refers to obtaining key-value pairs from the different components. 
The results show that the key-value pairs from the gaze autoencoder have a limited effect on the object detector, while the global-memory position knowledge of the object detector can optimize the performance of the gaze autoencoder.

\begin{figure}[!t]
  \centering
  \includegraphics[width=0.78\linewidth]{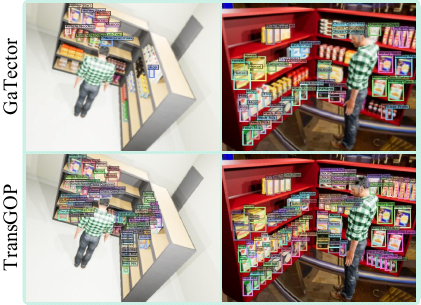}
  \caption{Object detection visualization of GaTector and TransGOP when IoU is 0.75.}
  \label{fig:vis_od}
\end{figure}

\begin{figure}[!t]
  \centering
  \includegraphics[width=0.78\linewidth]{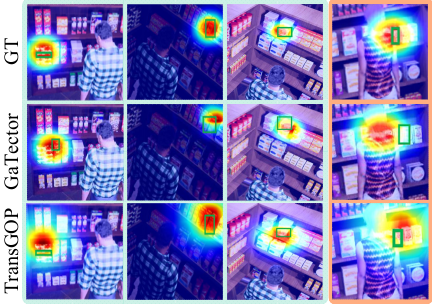}
  \caption{Gaze object prediction visualization results of GaTector and TransGOP.}
  \label{fig:vis}
\end{figure}

\section{Visualization}
\noindent\textbf{Visualization about object detection.}
Fig.~\ref{fig:vis_od} shows the qualitative results of object detection for GaTector and our proposed TransGOP when IoU is 0.75. Our TransGOP outperforms GaTector in detecting objects located at the intersection of people and goods or the edge of the shelf.

\noindent\textbf{Visualization about gaze object prediction.}
As shown in Fig.~\ref{fig:vis}, the GaTector predicts a relatively accurate heatmap, but due to its poor object detection performance, it cannot accurately locate the gaze box. In contrast, scene our TransGOP can learn precise location information from the object detector, improving the accuracy of gaze object prediction. The last column is failure cases.

\section{Conclusion}
\label{sec:conclusion}
This paper proposes an end-to-end Transformer-based gaze object prediction method named TransGOP, which consists of an object detector and a gaze regressor. The object detector is a DETR-like method to predict object location and category. Transformer-based gaze autoencoder is designed to build the long-range human gaze relationship in the gaze regressor. 
Meanwhile, to improve the regression of gaze heatmap, we propose an object-to-gaze cross-attention mechanism that utilizes the key-value pairs from the object detector as the input of the cross-attention block in the decoder layer of gaze autoencoder, so can help the model to predict more accurate gaze heatmap. 
Moreover, we also propose a novel gaze box loss that only focuses on the gaze energy in the gaze box to jointly optimize the performance of the object detector and gaze regressor. 
Finally, comprehensive experiments on the GOO-Synth and GOO-Real dataset demonstrates the effectiveness of our TransGOP.

\bigskip

\bibliography{aaai24}

\end{document}